\documentclass[11pt]{article}

\usepackage[utf8]{inputenc}
\usepackage[T1]{fontenc}
\usepackage{mathptmx}  
\usepackage[margin=1in]{geometry}

\usepackage{amsmath,amsfonts,amssymb,bm}

\usepackage{graphicx}
\usepackage{subcaption}
\usepackage{booktabs}

\usepackage{hyperref}
\usepackage{url}
\usepackage{natbib}

\title{On Catastrophic Forgetting in Low-Rank Decomposition-Based Parameter-Efficient Fine-Tuning}

\author{
Muhammad Ahmad\textsuperscript{1}, \quad
Jingjing Zheng\textsuperscript{2}, \quad
Yankai Cao\textsuperscript{3,*} \\[1ex]
\textsuperscript{1}Department of Electrical and Computer Engineering, The University of British Columbia \\
\textsuperscript{2}Department of Mathematics, The University of British Columbia \\
\textsuperscript{3}Department of Chemical and Biological Engineering, The University of British Columbia \\[1ex]
\texttt{sahmadmu@student.ubc.ca}, \quad
\texttt{jjzheng233@gmail.com}, \quad
\texttt{yankai.cao@ubc.ca} \\[1ex]
\textsuperscript{*}Corresponding author
}

\date{}

\begin{document}

\maketitle

\begin{abstract}
Parameter-efficient fine-tuning (PEFT) based on low-rank decomposition, such as LoRA, has become a standard for adapting large pretrained models. However, its behavior in sequential learning—specifically regarding catastrophic forgetting—remains insufficiently understood. In this work, we present an empirical study showing that forgetting is strongly influenced by the geometry and parameterization of the update subspace. While methods that restrict updates to small, shared matrix subspaces often suffer from task interference, tensor-based decompositions (e.g., LoRETTA) mitigate forgetting by capturing richer structural information within ultra-compact budgets, and structurally aligned parameterizations (e.g., WeGeFT) preserve pretrained representations. Our findings highlight update subspace design as a key factor in continual learning and offer practical guidance for selecting efficient adaptation strategies in sequential settings.
\end{abstract}

\section{Introduction}
\label{sec:intro}

Parameter-efficient fine-tuning (PEFT) has become a widely adopted paradigm for adapting large pretrained models to downstream tasks, as it drastically reduces training and storage costs by updating only a small subset of parameters while keeping the backbone frozen. Building on the success of Low-Rank Adaptation (LoRA) \cite{hu2021lora}, a large number of low-rank decomposition–based PEFT methods have emerged in the past few years, including adaptive rank allocation \cite{zhang2023adalora}, improved initialization schemes (PiSSA and LoRA-GA) \cite{meng2024pissa,wang2024lora}, low-rank plus parameter-sharing designs (WeGeFT) \cite{savadikar2025wegeft}, tensor-decomposition–based adapters \cite{yang2024loretta,bershatsky2024lotr}, and multi-subspace-based method \cite{AdaMSS2025_Zheng}. These methods all share the common principle of constraining parameter updates to a low-dimensional subspace while retaining strong downstream performance.

However, in many practical applications, models are required to learn from a sequence of tasks or to handle multiple tasks jointly, rather than being fine-tuned once and deployed in isolation \cite{verwimp2023continual}. In such sequential or multi-task learning scenarios, catastrophic forgetting, where performance on previously learned tasks degrades as new tasks are trained, becomes a central challenge \cite{kirkpatrick2017overcoming,MCCLOSKEY1989109}. While catastrophic forgetting has been extensively studied, much less is known about how re-parameterization–based fine-tuning behaves in these settings. In particular, it remains unclear whether constraining updates to low-dimensional subspaces alleviates task interference, and how different re-parameterization designs impact knowledge retention across tasks.

To bridge this gap, we conduct a controlled empirical study of catastrophic forgetting in low-rank decomposition–based efficient fine-tuning. We systematically compare a range of representative methods, including LoRA \cite{hu2021lora}, PiSSA \cite{meng2024pissa}, LoRETTA \cite{yang2024loretta}, and WeGeFT \cite{savadikar2025wegeft}, under both sequential and multi-task learning protocols. By evaluating their behavior across multiple vision benchmarks, we aim to answer the following questions: (i) how does constraining the update subspace affect forgetting compared to full fine-tuning? (ii) how do different low-rank decomposition-based designs influence task interference and knowledge retention? and (iii) what design principles lead to more robust PEFT under continual learning? Our results reveal a clear connection between the geometry of the update subspace and forgetting behavior, offering new insights into the trade-off between parameter efficiency and continual-learning stability.

\section{Methods}

We conduct experiments using a Vision Transformer (ViT) pretrained on ImageNet-1K as the base model to investigate how the choice of Parameter-Efficient Fine-Tuning (PEFT) method influences catastrophic forgetting in a continual learning setting. During fine-tuning, all backbone parameters are frozen, and only the parameters introduced by the PEFT modules are updated, with hyperparameters for each method selected based on validation performance. We adopt a sequential training protocol in which the pretrained ViT is fine-tuned on a series of four image classification tasks. After completing each task, the model is evaluated on all previously seen tasks to assess knowledge retention and quantify forgetting over time.

\subsection{Low-rank decomposition-based Fine-tuning}

In addition to full fine-tuning (FF), we compare four state-of-the-art PEFT methods that differ in how they adapt pretrained parameters while maintaining efficiency and stability during continual learning. Each method introduces unique mechanisms to balance adaptability and memory retention:

\begin{itemize}
\item \textbf{LoRA} \cite{hu2021lora} – Low-Rank Adaptation decomposes weight updates into rank-constrained matrices that are trained while keeping the original pretrained weights frozen.  
\item \textbf{PiSSA} \cite{meng2024pissa} – Principal Singular Values and Singular Vectors Adaptation updates only the principal components of pretrained weights. 
\item \textbf{LoRETTA} \cite{yang2024loretta} – LoRETTA employs tensor-train decomposition for ultra-parameter-efficient fine-tuning, featuring tensorized adapters and small tensor factor weight parameterization to cut trainable parameters.
\item \textbf{WeGeFT} \cite{savadikar2025wegeft} – WeGeFT constrains updates to a subspace aligned with pretrained weights and shares parameters across layers to substantially reduce the number of trainable parameters.
\end{itemize}

\subsection{Tasks}

The continual learning benchmark consists of four sequential tasks drawn from diverse visual domains. The dataset will be divided into training, validation and testing sets. Each task is designed to require distinct feature representations while remaining within the scope of image classification:

\begin{itemize}
    \item \textbf{Bird Species Classification} – Recognition of different bird species using the CUB-200-2011 dataset \cite{ezclap_bird_species}.
    \item \textbf{Land Use Classification} – Categorization of geospatial imagery into different land-use types using the EuroSat dataset \cite{helber2019eurosat}.
    \item \textbf{Natural Scene Recognition} – Classification of natural scenes sourced from the Intel image classification dataset \cite{intel_image_classification}.
    \item \textbf{Sports Image Classification} – Recognition of sports categories in images from the sports image classification dataset \cite{vieanh_sports_classification}.
\end{itemize}

This task sequence introduces progressive domain shifts—from fine-grained recognition to scene-level understanding—forming a robust benchmark for evaluating forgetting dynamics.

\begin{table*}[t]
\centering
\caption{Comparison of performance and parameter efficiency across different fine-tuning methods on ViT-Base and ViT-Large. The configurations for each method are: LoRA ($r{=}8$), PiSSA ($r{=}8$), LoRETTA ($R{=}5$), and WeGeFT ($r{=}16$).}
\label{tab:results}

\begin{subtable}{0.48\linewidth}
    \centering
    \caption{ViT-Base}
    \resizebox{\linewidth}{!}{
    \begin{tabular}{lccccc}
    \toprule
    \textbf{Metric} & \textbf{FF} & \textbf{LoRA} & \textbf{PiSSA} & \textbf{LoRETTA} & \textbf{WeGeFT} \\
    \midrule
    \# Trainable Params & 85.8M & 313K & 313K & 57K & 54K \\
    Avg. Forgetting & 0.0740 & 0.1491 & 0.2089 &  0.0717  & 0.0752 \\
    Final Accuracy (\%) & 92.70 & 86.79 & 81.25 &  92.65  & 92.30 \\
    \bottomrule
    \end{tabular}
    }
\end{subtable}
\hfill 
\begin{subtable}{0.48\linewidth}
    \centering
    \caption{ViT-Large}
    \resizebox{\linewidth}{!}{
    \begin{tabular}{lccccc}
    \toprule
    \textbf{Metric} & \textbf{FF} & \textbf{LoRA} & \textbf{PiSSA} & \textbf{LoRETTA} & \textbf{WeGeFT} \\
    \midrule
    \# Trainable Params & 303.3M & 835K & 835K & 132K & 65K \\
    Avg. Forgetting & 0.0685 & 0.0228 & 0.2339 & 0.0338 & 0.0294 \\
    Final Accuracy (\%) & 93.20 & 96.65 & 80.72 & 95.52 &  95.97 \\
    \bottomrule
    \end{tabular}
    }
\end{subtable}

\end{table*}

\subsection{Metrics}

After training on each task, the model is evaluated on all previously encountered tasks. To quantify forgetting, we employ the metric proposed by Chaudhry et al. \cite{chaudhry2018riemannian}, which measures the performance degradation on previous tasks after learning subsequent tasks.

For a given task $j$, forgetting is defined as:
\begin{equation}
F_j = \max_{l \in \{1, \dots, N-1\}} A_{l,j} - A_{N,j}
\end{equation}
where $A_{l,j}$ denotes the accuracy on task $j$ after completing training on task $l$, and $A_{N,j}$ is the accuracy on task $j$ after training on all $N$ tasks. The overall average forgetting across tasks is then computed as:
\begin{equation}
\bar{F} = \frac{1}{N - 1} \sum_{j=1}^{N-1} F_j
\end{equation}
In addition to forgetting, we use per-task accuracy and average incremental accuracy after each training stage to capture overall model adaptability and retention trends. The final average accuracy is defined as
\[
\mathrm{ACC}_{\mathrm{final}} = \frac{1}{N}\sum_{i=1}^{N} A_{i,N}.
\]

\section{Experimental Results and Analysis}
 
\begin{figure}[t]
    \centering
    \begin{subfigure}[t]{0.32\linewidth}
        \centering
        \includegraphics[width=\linewidth]{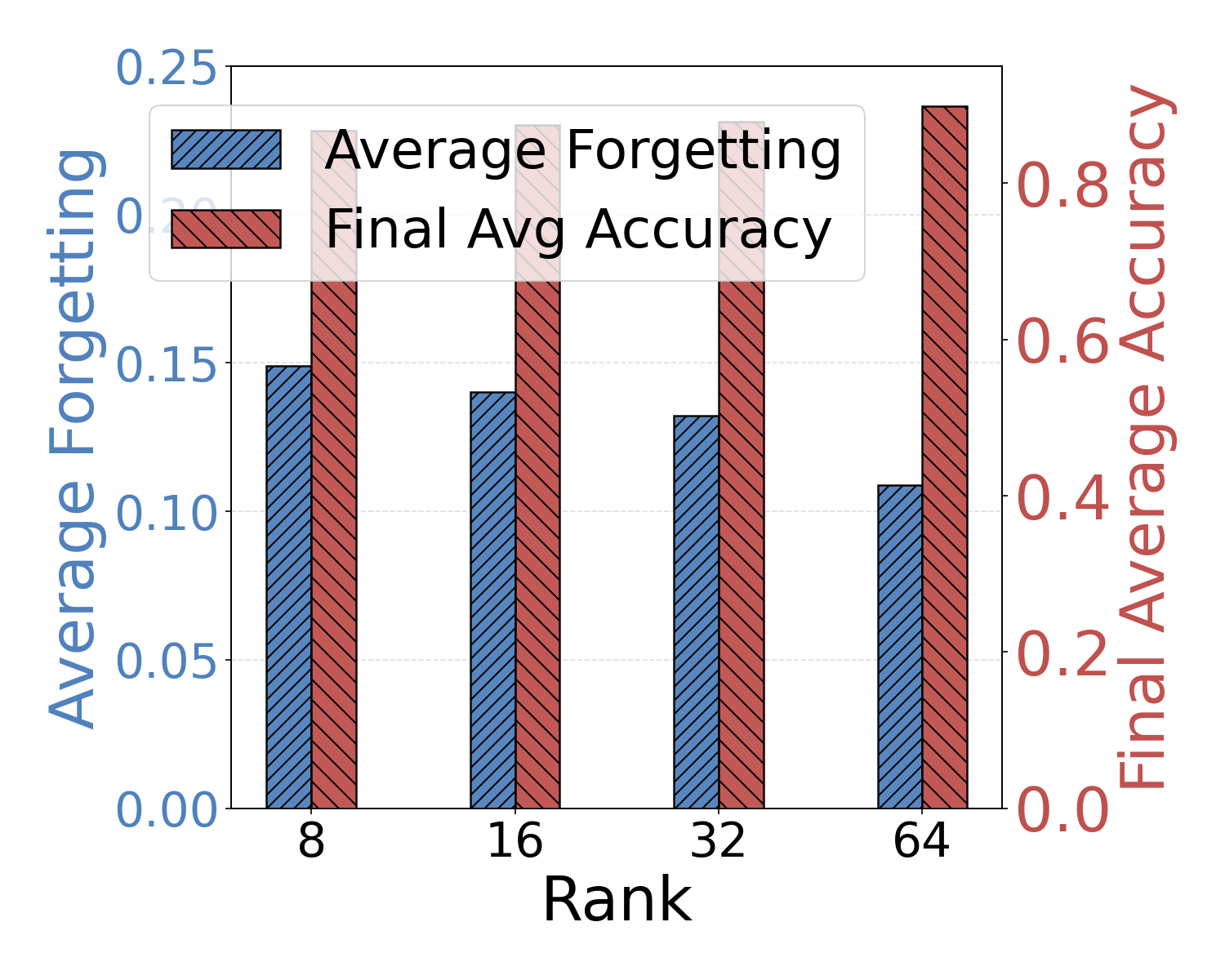}
        \caption{LoRA}
    \end{subfigure}
    \hfill
    \begin{subfigure}[t]{0.32\linewidth}
        \centering \includegraphics[width=\linewidth]{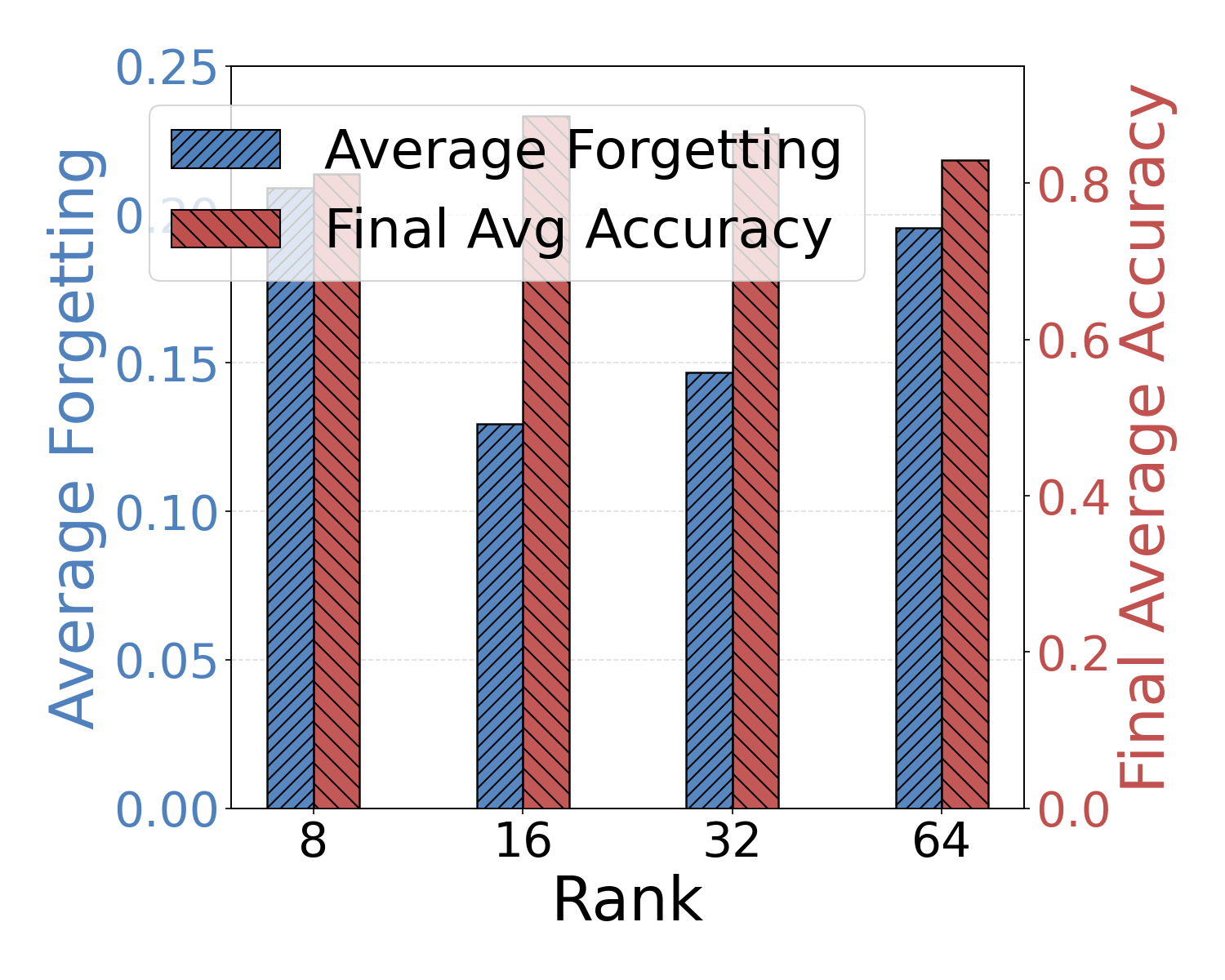}
        \caption{PiSSA}
    \end{subfigure}
 \hfill
    \begin{subfigure}[t]{0.32\linewidth}
        \centering \includegraphics[width=\linewidth]{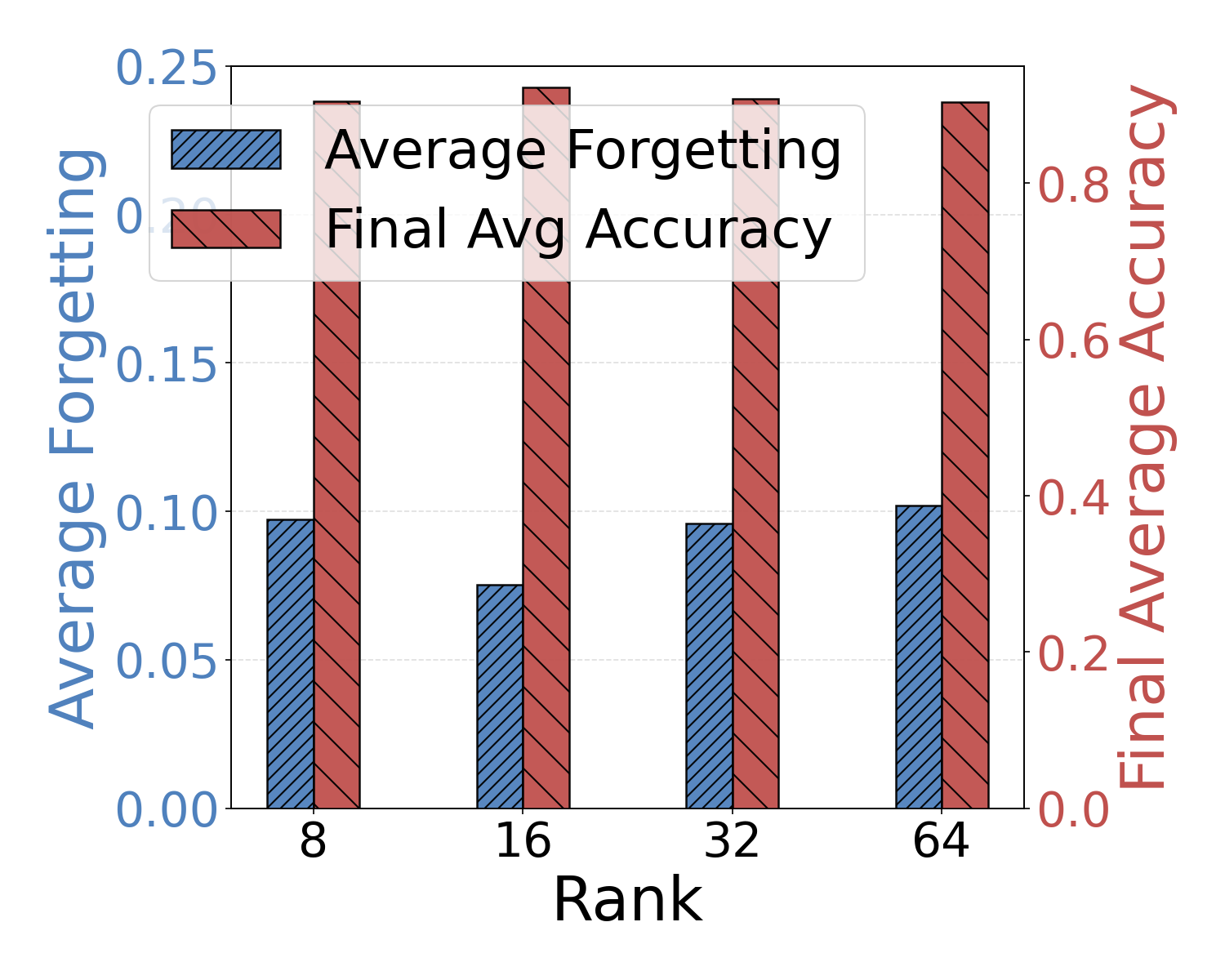}
        \caption{WeGeFT}
    \end{subfigure}

    \caption{Performance comparison of different fine-tuning methods on ViT-Base under varying update space dimensions (rank).}
    \label{fig:multi_tasks}
\end{figure}

\begin{figure}[t]
    \centering
    \begin{subfigure}[t]{0.32\linewidth}
        \centering
        \includegraphics[width=\linewidth]{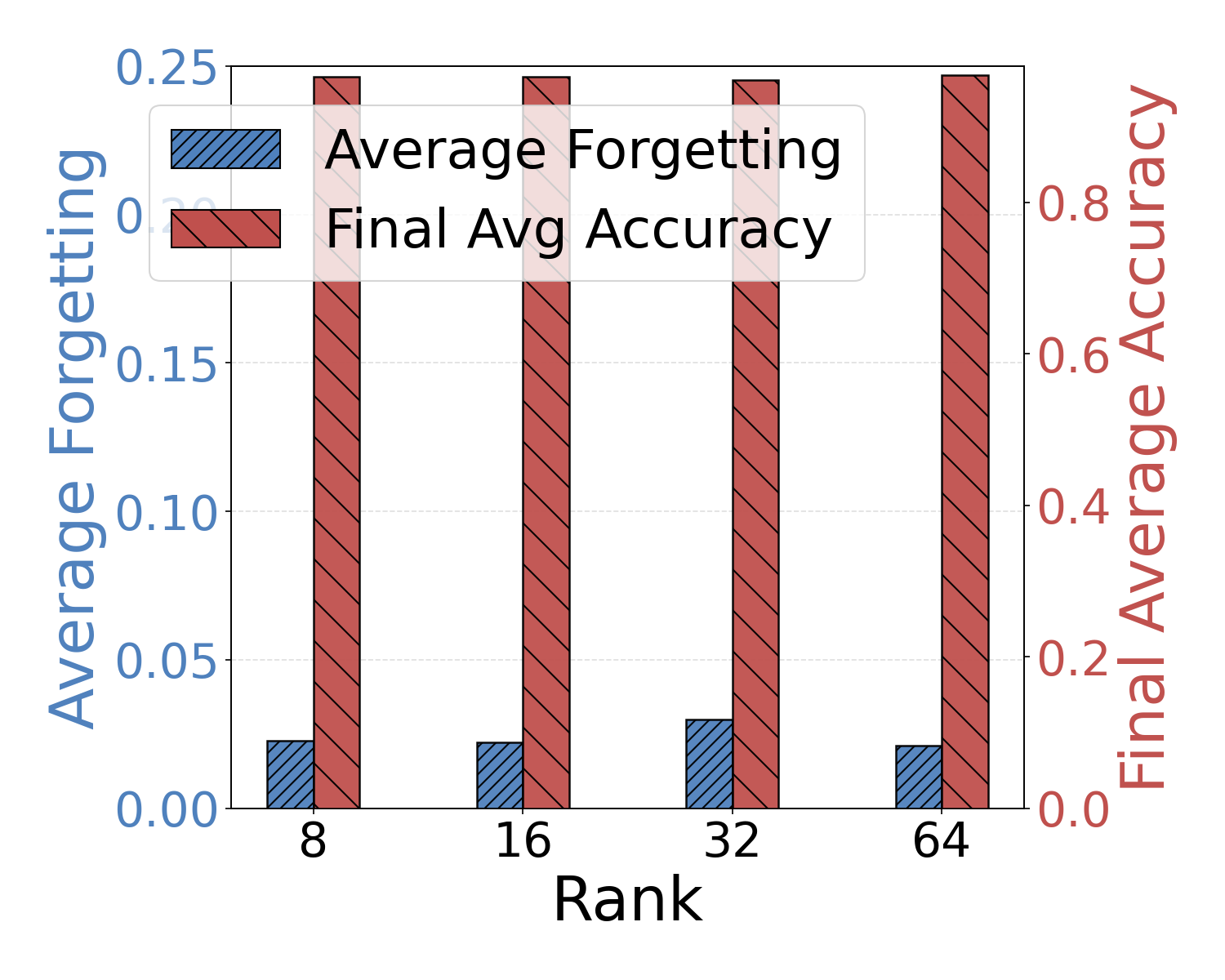}
        \caption{LoRA}
    \end{subfigure}
    \hfill
    \begin{subfigure}[t]{0.32\linewidth}
        \centering \includegraphics[width=\linewidth]{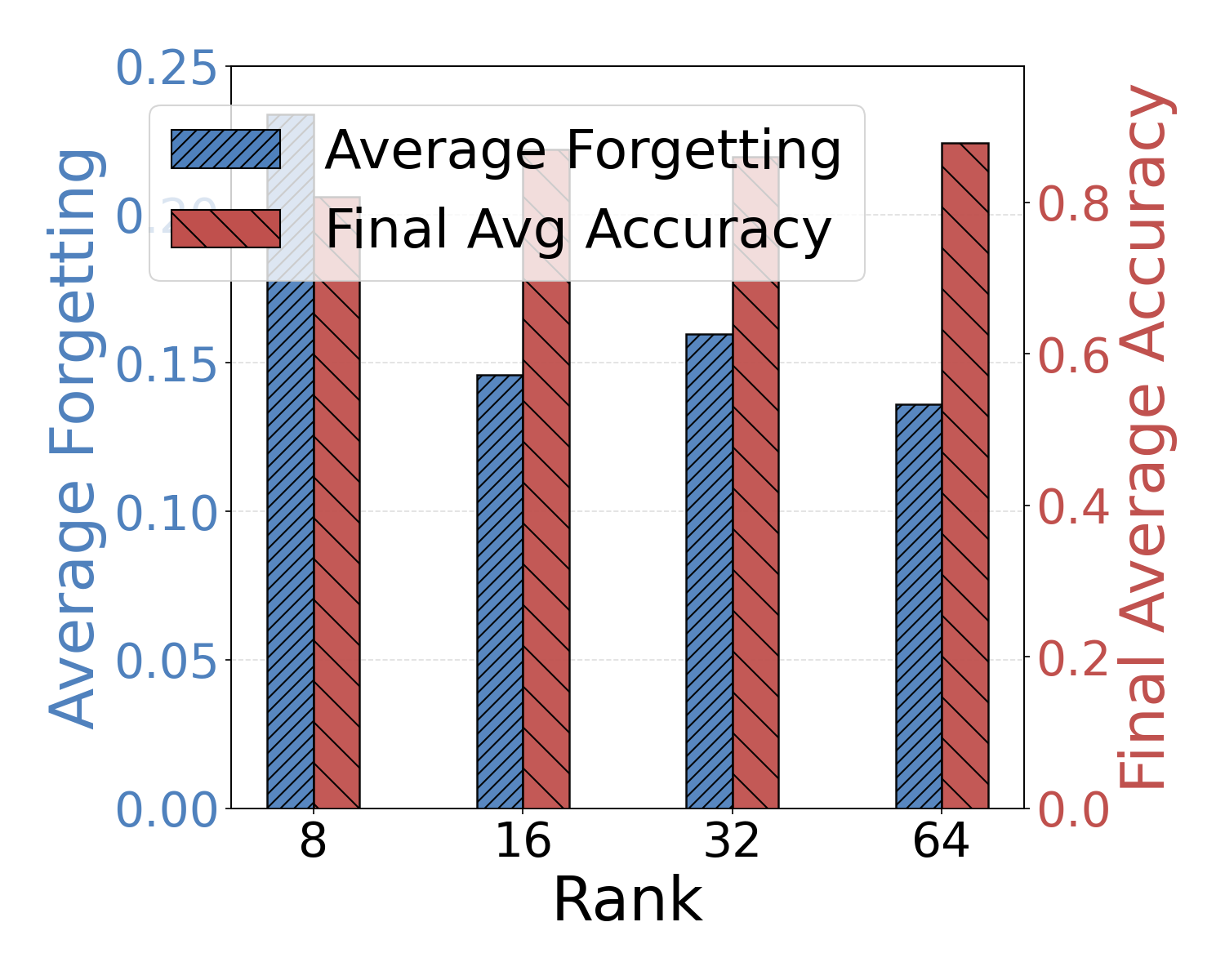}
        \caption{PiSSA}
    \end{subfigure}
 \hfill
    \begin{subfigure}[t]{0.32\linewidth}
        \centering \includegraphics[width=\linewidth]{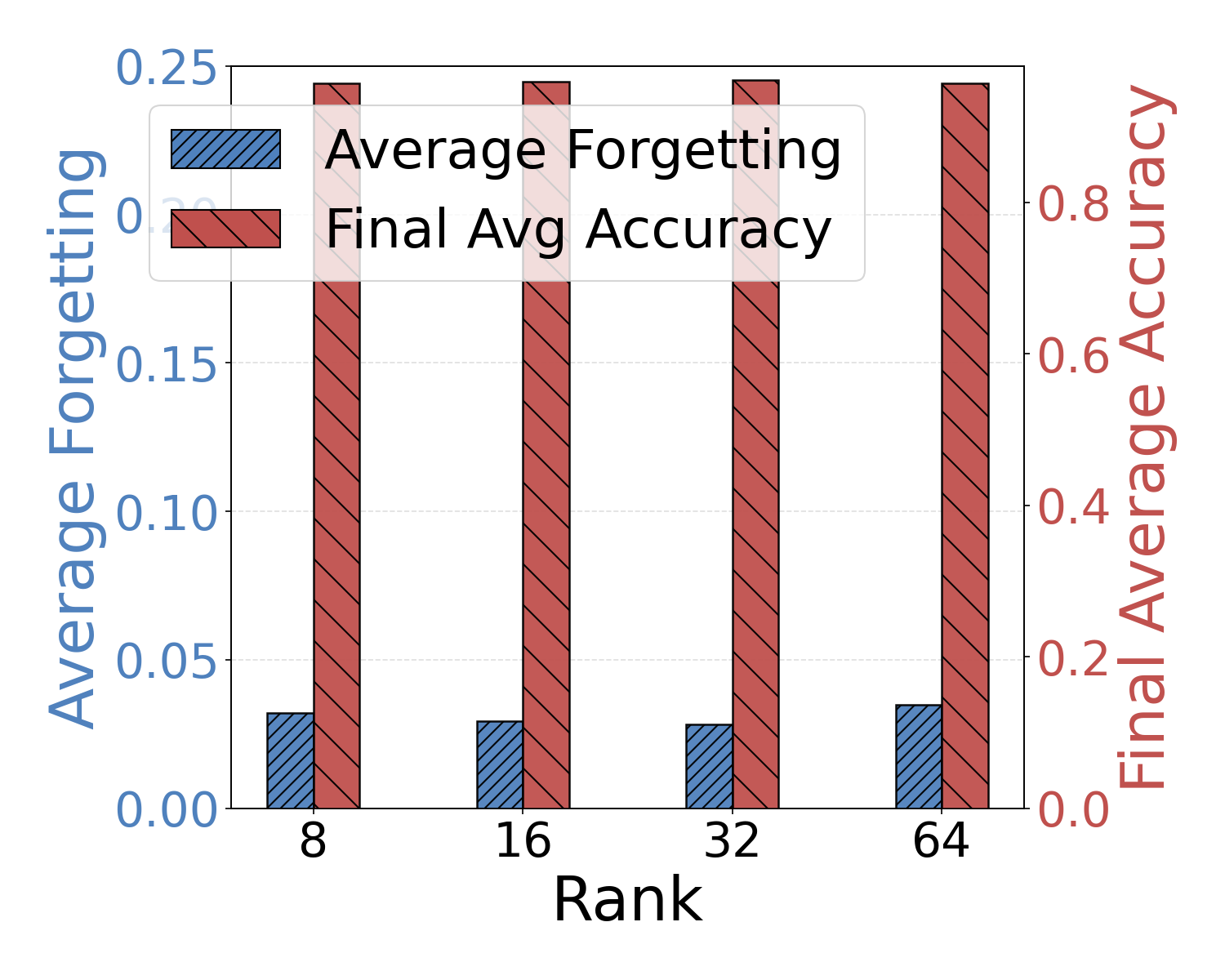}
        \caption{WeGeFT}
    \end{subfigure}

    \caption{Performance comparison of different fine-tuning methods on ViT-Large under varying update space dimensions (rank).}
    \label{fig:multi_tasks-large}
\end{figure}

Table~\ref{tab:results} and Figs.~\ref{fig:multi_tasks}-\ref{fig:multi_tasks-large} report the results of full fine-tuning (FF) and several low-rank decomposition–based PEFT methods under a sequential continual learning setting. The empirical results reveal a clear trend emerges: the extent of catastrophic forgetting is closely related to how much flexibility the method allows in parameter updates. Full fine-tuning represents an extreme case in which updates are unconstrained and span the entire parameter space. This flexibility allows updates from different tasks to occupy distinct directions, which helps reduce interference between tasks. 

In contrast, LoRA explicitly constrains updates to a low-dimensional subspace. Under small ranks, updates from different tasks are forced to share a limited set of directions, which is reflected in higher forgetting. As the rank increases, forgetting decreases steadily, indicating that a less restrictive update space is associated with improved knowledge retention. This monotonic trend is consistently observed across different ranks, as shown in Fig.~\ref{fig:multi_tasks} (a).

A more pronounced effect is observed for PiSSA. PiSSA restricts updates strictly to the principal singular subspace of the pretrained weights, whose directions typically encode highly general and cross-task–shared representations. At low ranks, this results in the most severe forgetting among all evaluated methods, and consistently exhibits higher forgetting than LoRA at comparable ranks, as shown in Figs.~\ref{fig:multi_tasks}-\ref{fig:multi_tasks-large}. Empirically, this behavior is associated with the fact that PiSSA enforces updates within the principal singular subspace of the pretrained weights, disrupting the pretrained weight space.

Different from LoRA and PiSSA, updates of WeGeFT are not arbitrary low-rank modifications but are constrained to lie within the subspace spanned by the pretrained weights. Empirically, this design enables WeGeFT to achieve consistently low forgetting across a wide range of ranks, even under strict parameter budgets, while maintaining competitive final average accuracy. Similarly, LoRETTA introduces a paradigm shift by departing from the conventional matrix decomposition framework entirely. By reshaping weights into high-dimensional tensors and employing Tensor Train (TT) decomposition, LoRETTA captures richer structural information and dependencies than simple low-rank matrices. This allows it to achieve competitive accuracy and minimal forgetting even under extreme parameter compression, demonstrating that efficient parameterization is as critical as subspace alignment.

Overall, the experimental results highlight two complementary empirical regimes for mitigating catastrophic forgetting in sequential learning. One regime is to allow sufficient update flexibility, as observed in higher-rank LoRA variants, or highly compressed methods like LoRETTA that capture rich information with minimal parameters via tensor decomposition. These approaches are consistently associated with lower forgetting. The other is to impose structured constraints on updates that maintain alignment with pretrained representations, as exemplified by WeGeFT. In contrast, methods that both strongly restrict updates to a small and highly shared subspace and disrupt the pretrained representation subspace, such as PiSSA at low ranks, consistently exhibit more severe forgetting.

\section{Conclusion}

This work investigates catastrophic forgetting in sequential continual learning under parameter-efficient fine-tuning (PEFT) methods. From the perspective of update subspaces, we show that forgetting behavior is closely related to the degree of update freedom allowed during adaptation. By comparing different decomposition designs, we further find that the parameterization of the update subspace, whether through low-rank matrices or high-dimensional tensors, plays a critical role in knowledge retention. Within this framework, LoRETTA demonstrates that tensor-based compression can capture rich structural information to minimize forgetting even with extreme parameter constraints, while WeGeFT serves as a representative method that learns the update subspace through a weighting mechanism to preserve the pretrained space. Together, these methods show that achieving a favorable balance between new-task adaptation and historical knowledge retention requires specific structural designs. These findings highlight the importance of jointly considering the size, geometry, and structural efficiency of update subspaces when designing PEFT methods for continual learning.

\bibliographystyle{plainnat}
\bibliography{references}

@article{helber2019eurosat,
  author    = {P. Helber and B. Bischke and A. Dengel and D. Borth},
  title     = {{EuroSAT}: A Novel Dataset and Deep Learning Benchmark for Land Use and Land Cover Classification},
  journal   = {IEEE Journal of Selected Topics in Applied Earth Observations and Remote Sensing},
  volume    = {12},
  number    = {7},
  pages     = {2217--2226},
  year      = {2019},
  note      = {[Online]. Available: \url{https://huggingface.co/datasets/timm/eurosat-rgb}}
}

@misc{intel_image_classification,
  author    = {{Analytics Vidhya} and {Intel}},
  title     = {{Intel Image Classification Challenge}},
  year      = {2018},
  howpublished = {Kaggle},
  note      = {[Online]. Available: \url{https://huggingface.co/datasets/miladfa7/Intel-Image-Classification}}
}

@misc{ezclap_bird_species,
  author    = {{Ez-Clap}},
  title     = {{Bird Species Dataset}},
  year      = {2023},
  howpublished = {Hugging Face},
  note      = {[Online]. Available: \url{https://huggingface.co/datasets/Ez-Clap/bird-species}}
}

@misc{vieanh_sports_classification,
  author    = {{Vieanh}},
  title     = {{Sports Image Classification Dataset}},
  year      = {2023},
  howpublished = {Hugging Face},
  note      = {[Online]. Available: \url{https://huggingface.co/datasets/vieanh/sports_img_classification}}
}

@inproceedings{chaudhry2018riemannian,
  title={Riemannian Walk for Incremental Learning: Understanding Forgetting and Intransigence},
  author={Chaudhry, Arslan and Dokania, Puneet K and Ajanthan, Thalaiyasingam and Torr, Philip HS},
  booktitle={Proceedings of the European Conference on Computer Vision (ECCV)},
  pages={532--547},
  year={2018}
}

@article{hu2021lora,
  title={LoRA: Low-Rank Adaptation of Large Language Models},
  author={Hu, Edward J and Shen, Yelong and Wallis, Phillip and Allen-Zhu, Zeyuan and Li, Yuanzhi and Wang, Shean and Wang, Lu and Chen, Weizhu},
  journal={arXiv preprint arXiv:2106.09685},
  year={2021}
}

@inproceedings{zhang2023adalora,
  title={Adaptive Budget Allocation for Parameter-Efficient Fine-Tuning},
  author={Zhang, Qingru and Chen, Minshuo and Bukharin, Alexander and He, Pengcheng and Cheng, Yu and Chen, Weizhu and Zhao, Tuo},
  booktitle={International Conference on Learning Representations},
  year={2023}
}

@inproceedings{meng2024pissa,
  title={PiSSA: Principal Singular Values and Singular Vectors Adaptation of Large Language Models},
  author={Meng, Fanxu and Wang, Zhaohui and Zhang, Muhan},
  booktitle={Neural Information Processing Systems},
  year={2024}
}

@inproceedings{wang2024lora,
  title={LoRA-GA: Low-Rank Adaptation with Gradient Approximation},
  author={Wang, Shaowen and Yu, Linxi and Li, Jian},
  booktitle={Neural Information Processing Systems},
  year={2024}
}

@article{bershatsky2024lotr,
  title={LoTR: Low Tensor Rank Weight Adaptation},
  author={Bershatsky, Daniel and Cherniuk, Daria and Daulbaev, Talgat and Oseledets, Ivan},
  journal={arXiv preprint arXiv:2402.01376},
  year={2024}
}

@inproceedings{yang2024loretta,
  title={LoRETTA: Low-Rank Economic Tensor-Train Adaptation for Ultra-Low-Parameter Fine-Tuning of Large Language Models},
  author={Yang, Yifan and Zhou, Jiajun and Wong, Ngai and Zhang, Zheng},
  booktitle={Proceedings of the 2024 Conference of the North American Chapter of the Association for Computational Linguistics: Human Language Technologies},
  pages={3161--3176},
  year={2024}
}

@inproceedings{savadikar2025wegeft,
  title={WeGeFT: Weight-Generative Fine-Tuning for Multi-Faceted Efficient Adaptation of Large Models},
  author={Savadikar, Chinmay and Song, Xi and Wu, Tianfu},
  booktitle={Forty-second International Conference on Machine Learning},
  year={2025},
  url={https://openreview.net/forum?id=K0sv5T2usb}
}

@inproceedings{AdaMSS2025_Zheng,
  author = {Zheng, Jingjing and Lu, Wanglong and Dong, Yiming and Ji, Chaojie and Cao, Yankai and Lin, Zhouchen},
  booktitle = {Advances in Neural Information Processing Systems},
  title = {AdaMSS: Adaptive Multi-Subspace Approach for Parameter-Efficient Fine-Tuning},
  year = {2025}
}

@incollection{MCCLOSKEY1989109,
  title = {Catastrophic Interference in Connectionist Networks: The Sequential Learning Problem},
  editor = {Gordon H. Bower},
  series = {Psychology of Learning and Motivation},
  publisher = {Academic Press},
  volume = {24},
  pages = {109-165},
  year = {1989},
  issn = {0079-7421},
  doi = {https://doi.org/10.1016/S0079-7421(08)60536-8}, 
  author = {Michael McCloskey and Neal J. Cohen}
}

@article{kirkpatrick2017overcoming,
  title={Overcoming Catastrophic Forgetting in Neural Networks},
  author={Kirkpatrick, James and Pascanu, Razvan and Rabinowitz, Neil and Veness, Joel and Desjardins, Guillaume and Rusu, Andrei A and Milan, Kieran and Quan, John and Ramalho, Tiago and Grabska-Barwinska, Agnieszka and others},
  journal={Proceedings of the National Academy of Sciences},
  volume={114},
  number={13},
  pages={3521--3526},
  year={2017},
  publisher={National Academy of Sciences}
}

@article{verwimp2023continual,
  title={Continual Learning: Applications and the Road Forward},
  author={Verwimp, Eli and Aljundi, Rahaf and Ben-David, Shai and Bethge, Matthias and Cossu, Andrea and Gepperth, Alexander and Hayes, Tyler L and H{\"u}llermeier, Eyke and Kanan, Christopher and Kudithipudi, Dhireesha and others},
  journal={arXiv preprint arXiv:2311.11908},
  year={2023}
}

@inproceedings{houlsby2019parameter,
  title={Parameter-Efficient Transfer Learning for NLP},
  author={Houlsby, Neil and Giurgiu, Andrei and Jastrzebski, Stanislaw and Morrone, Bruna and De Laroussilhe, Quentin and Gesmundo, Andrea and Attariyan, Mona and Gelly, Sylvain},
  booktitle={International Conference on Machine Learning},
  pages={2790--2799},
  year={2019},
  organization={PMLR}
}

@inproceedings{wang2022learning,
  title={Learning to Prompt for Continual Learning},
  author={Wang, Zifeng and Zhang, Zizhao and Lee, Chen-Yu and Zhang, Han and Sun, Ruoxi and Ren, Xiaoqi and Su, Guolong and Perot, Vincent and Dy, Jennifer and Pfister, Tomas},
  booktitle={Proceedings of the IEEE/CVF Conference on Computer Vision and Pattern Recognition},
  pages={139--149},
  year={2022}
}

@inproceedings{wang2022dualprompt,
  title={DualPrompt: Complementary Prompting for Rehearsal-free Continual Learning},
  author={Wang, Zifeng and Zhang, Zizhao and Ebrahimi, Sayna and Sun, Ruoxi and Zhang, Han and Lee, Chen-Yu and Ren, Xiaoqi and Su, Guolong and Perot, Vincent and Dy, Jennifer and others},
  booktitle={European Conference on Computer Vision},
  pages={631--648},
  year={2022},
  organization={Springer}
}

@inproceedings{yu2024boosting,
  title={Boosting Continual Learning of Vision-Language Models via Mixture-of-Experts Adapters},
  author={Yu, Jiazuo and Zhuge, Yunzhi and Zhang, Lu and Hu, Ping and Wang, Dong and Lu, Huchuan and He, You},
  booktitle={Proceedings of the IEEE/CVF Conference on Computer Vision and Pattern Recognition},
  pages={23219--23230},
  year={2024}
}

@inproceedings{huang2024expand,
  title={Expand and Merge: Continual Learning with the Guidance of Fixed Text Embedding Space},
  author={Huang, Yujun and Zhang, Wentao and Wang, Ruixuan},
  booktitle={2024 International Joint Conference on Neural Networks (IJCNN)},
  pages={1--8},
  year={2024},
  organization={IEEE}
}

@article{li2024atlas,
  title={Atlas: Adapter-based Multi-modal Continual Learning with a Two-Stage Learning Strategy},
  author={Li, Hong and Tan, Zhiquan and Li, Xingyu and Huang, Weiran},
  journal={arXiv preprint arXiv:2410.10923},
  year={2024}
}

@article{coleman2025parameter,
  title={Parameter-Efficient Continual Fine-Tuning: A Survey},
  author={Coleman, Eric Nuertey and Quarantiello, Luigi and Liu, Ziyue and Yang, Qinwen and Mukherjee, Samrat and Hurtado, Julio and Lomonaco, Vincenzo},
  journal={arXiv preprint arXiv:2504.13822},
  year={2025}
}

@inproceedings{serra2018overcoming,
  title={Overcoming Catastrophic Forgetting with Hard Attention to the Task},
  author={Serra, Joan and Suris, Didac and Miron, Marius and Karatzoglou, Alexandros},
  booktitle={International Conference on Machine Learning},
  pages={4548--4557},
  year={2018},
  organization={PMLR}
}

@article{biderman2024lora,
  title={LoRA Learns Less and Forgets Less},
  author={Biderman, Dan and Portes, Jacob and Ortiz, Jose Javier Gonzalez and Paul, Mansheej and Greengard, Philip and Jennings, Connor and King, Daniel and Havens, Sam and Chiley, Vitaliy and Frankle, Jonathan and others},
  journal={arXiv preprint arXiv:2405.09673},
  year={2024}
}

\appendix
\section{Hyper-parameter Configuration}\label{Hyper-parameter}

To ensure a fair comparison across methods, the batch size, optimizer, and number of epochs were kept consistent for all fine-tuning methods. The learning rate for each method was selected based on the value that yielded the highest Averaged Anytime Accuracy. The complete set of hyper-parameters used in the experiments is presented in Table~\ref{tab:hyperparams}.

\begin{table}[h]
\centering
\caption{Hyperparameters used for full fine-tuning (FF) and parameter-efficient methods.}
\begin{tabular}{l|ccccc}
\hline
\textbf{Hyperparameter} & \textbf{FF} & \textbf{LoRA} & \textbf{PiSSA} & \textbf{LoRETTA} & \textbf{WeGeFT} \\
\hline
Optimizer & \multicolumn{5}{c}{Adam} \\
Number of epochs per task & \multicolumn{5}{c}{10} \\
Batch size & \multicolumn{5}{c}{15} \\
\hline
Learning rate & 0.0001 & 0.001 & 0.0001 & 0.01 & 0.001 \\
\hline
\end{tabular}
\label{tab:hyperparams}
\end{table}

\section{Related Work}

Recent progress at the intersection of parameter-efficient fine-tuning (PEFT) and continual learning (CL) points toward a viable pathway for building AI systems that are both adaptive and efficient \cite{coleman2025parameter}. Empirical evidence further suggests that constraining parameter updates can, in many settings, mitigate forgetting relative to full fine-tuning \cite{serra2018overcoming,biderman2024lora}. 
Task-specific adapter–based methods, for example, often exhibit stronger task isolation due to their explicit parameter separation \cite{houlsby2019parameter,wang2022learning,wang2022dualprompt,yu2024boosting}. However, completely avoiding parameter sharing across tasks typically leads to a number of trainable parameters that grows linearly with the task sequence, which runs counter to the efficiency goals of PEFT. As a result, adapter-based approaches, such as ATLAS \cite{li2024atlas} and Expand and Merge \cite{huang2024expand}, are commonly designed to operate under an explicit trade-off between parameter sharing and knowledge retention, reflecting practical constraints on efficiency and scalability in continual learning settings.

\end{document}